\documentclass[letterpaper]{article} 
\usepackage{aaai2026}  
\usepackage{times}  
\usepackage{helvet}  
\usepackage{courier}  
\usepackage[hyphens]{url}  
\usepackage{graphicx} 
\usepackage{natbib}  
\usepackage{caption} 
\usepackage{graphicx}
\usepackage{caption}
\usepackage{subcaption}

\usepackage{algorithm}
\usepackage{algorithmic}

\usepackage{newfloat}
\usepackage{listings}
\DeclareCaptionStyle{ruled}{labelfont=normalfont,labelsep=colon,strut=off} 
\floatstyle{ruled}
\newfloat{listing}{tb}{lst}{}
\floatname{listing}{Listing}
\title{Amortized In-Context Mixed Effect Transformer Models: A Zero-Shot Approach for Pharmacokinetics}

\author {
    César A. Ojeda M., 
    Wilhelm Huisinga, 
    Purity Kavwele, 
    Ramsés J. Sánchez,
    Niklas Hartung
}
\affiliations {
    University of Potsdam \\
    \{ojedamarin, huisinga, purity.kavwele, niklas.hartung\}@uni-potsdam.de
}

\usepackage{bibentry}
\usepackage{amsfonts}
\usepackage{amsmath}
\usepackage{mathtools}
\usepackage[table]{xcolor}
\usepackage{booktabs}
\usepackage{graphicx}
\usepackage{adjustbox}
\usepackage{subcaption}
\usepackage[percent]{overpic}  

\begin{document}

\maketitle

\begin{abstract}
Accurate dose–response forecasting under sparse sampling is central to precision pharmacotherapy. 
We present the Amortized In-Context Mixed-Effect Transformer (AICMET) model, a transformer‑based latent‑variable framework that unifies mechanistic compartmental priors with amortized in‑context Bayesian inference. 
AICMET is pre‑trained on hundreds of thousands of synthetic pharmacokinetic trajectories with  Ornstein-Uhlenbeck priors over the parameters of compartment models, endowing the model with strong inductive biases and enabling zero‑shot adaptation to new compounds.
At inference time, the decoder conditions on the collective context of previously profiled trial participants, generating calibrated posterior predictions for newly enrolled patients after a few early drug concentration measurements. 
This capability collapses traditional model‑development cycles from weeks to hours while preserving some degree of expert modelling. 
Experiments across public datasets show that AICMET attains state‑of‑the‑art predictive accuracy and faithfully quantifies inter‑patient variability—outperforming both nonlinear mixed‑effects baselines and recent neural ODE variants. 
Our results highlight the feasibility of transformer‑based, population‑aware neural architectures as offering a new alternative for bespoke pharmacokinetic modeling pipelines, charting a path toward truly population-aware personalized dosing regimens.
\end{abstract}


\section{Introduction}

Recent advances in scalable deep‑learning techniques have brought the long‑standing vision of truly personalized therapeutics—often described as the “holy grail” of precision medicine—within practical reach, positioning it as a cornerstone of modern drug development and therapeutic use. Yet a key challenge remains markedly underexplored in contemporary neural architectures: the principled modeling of an ensemble of longitudinal, patient‑specific response trajectories as a function of administered doses and with few observations per individual.

Traditionally, pharmacokinetic (PK) modeling relies on systems of ordinary differential equations (ODEs) to describe the temporal evolution of drug concentrations within the body. 
In clinical trials, however, the available data are typically sparse—often consisting of only a handful of observations per individual. To compensate, classical approaches employ nonlinear mixed‑effects (NLME) models in which ODE parameters are decomposed into fixed effects (capturing population‑level kinetics) and random effects (accounting for inter‑individual variability). This hierarchical strategy enables the sharing of statistical strength across subjects while preserving subject‑specific inference, even under limited observation regimes \cite{lavielle2014}.

Despite their successes, these traditional pipelines require manual specification of drug‑specific kinetic systems—often crafted de novo for each compound—along with labor‑intensive calibration and training workflows. 
Some deep learning alternatives have been proposed for individual-level inference in rich data settings, such as neural ODEs \cite{lu2021deep}. However, the specific challenge of representing sparse hierarchical data, critical for robust population‑level inference, has not been addressed so far.

A modern framework should  retain the representational flexibility of neural ODEs while remaining data‑efficient and population‑aware. Specifically, it must accommodate sparse, irregularly sampled datasets and explicitly model latent study population structure, thereby enabling generalization across compounds, individuals and study design while adapting to patient‑specific observations.

In this work we leverage the success of transformer architectures for in‑context Bayesian inference \cite{mittal2025amortized, muller2021transformers, berghaus2024foundation, seifner2025zeroshot, seifner2025foundation} and translate these methodologies to mixed‑effects pharmacokinetic models. We train a transformer decoder that operates in context: it can both generate synthetic patient trajectories and, crucially, exploit the collective context provided by previously profiled trial participants to deliver dose‑conditioned predictions for newly enrolled patients receiving the same drug. The ability to generate synthetic patients for specific populations on a given drug allows us to characterize the inter‑patient variability required to optimize individualized dosing regimens, evaluate safety margins under covariate shifts, and quantify the predictive uncertainty needed for adaptive trial design and regulatory submission.

Similar to \citet{seifner2025zeroshot, seifner2025foundation} and \citet{berghaus2024foundation}, we first propose a statistical model that imposes stochastic differential equation priors on classical compartment ODE model parameters. Furthermore, we introduce a hidden‑variable formulation akin to neural processes \cite{garnelo2018neural, nguyen2022transformer}, but with a hierarchical structure: a global latent representation captures population‑level effects, while local latents model individual‑specific deviations. Our transformer based decoder integrates neural operator ideas \cite{lu2021learning,seifner2025zeroshot,seifner2025foundation} by encoding continuous time directly in the attention queries.

By pre‑training on a massive corpus of simulated PK trajectories, the model internalizes model-specific pharmacological inductive biases. Consequently, bringing a new compound online involves only a brief, sometimes zero‑shot, adaptation step---shrinking model development timelines from weeks to seconds and freeing domain experts to influence therapeutic decisions rather than hand‑crafting ODE systems.
In short, we provide:  (i) a unified probabilistic framework combining compartmental PK models with hierarchical latent-variable transformers to capture both population- and subject-level structures; (ii) an amortized in-context inference mechanism enabling calibrated predictions from as little as a single observation per individual; (iii) a dose-aware neural architecture that models continuous-time dynamics and naturally handles irregular sampling; and (iv) a pretrained generative model supporting zero-shot forecasting, safety margin estimation, and predictive uncertainty quantification.
Together, these contributions outline a data‑efficient, population‑aware, and mechanistically grounded pathway towards truly personalized pharmacotherapy.

\section{Related Work}
Machine learning solutions for NLME modelling span a wide range of methodologies, from traditional nonparametric Gaussian process regression methods \cite{shi2012mixed} to more scalable variants that condition the GP mean functions on neural networks \cite{chung2020deep}. Neural ODE-based approaches have also been applied in this context \cite{nazarovs2022mixed}, although they typically impose a linear dependence on the fixed effects and still rely on computationally expensive and often unstable adjoint methods for training. Specifically tailored solutions for pharmacokinetics \cite{lu2021deep} have demonstrated that neural ODEs can learn both complex latent dynamics and response times as a function of dosing schemes. However, these methods are generally trained on large datasets and ignore the hierarchical, population-level structure central to NLME modeling. Within pharmacology \cite{arruda2023amortized}, NLME parameters have been estimated using neural posterior estimation. This approach, however, is limited to parameter recovery and cannot discover the underlying dynamics. Our framework instead acts as a prior-fitted network over the dynamics, employing nonparametric priors directly on the training data.

Transformer-based models, on the other hand, have recently been explored for their capacity to perform Bayesian inference via in-context learning \cite{muller2021transformers, mittal2025amortized}. In particular, \cite{muller2021transformers} introduce the notion of prior-fitted networks, drawing a connection to simulation-based inference where domain-specific priors are incorporated, and generalization is achieved through zero-shot adaptation to sparse data. In contrast, \cite{mittal2025amortized, mittal2025context} present a comprehensive evaluation across training objectives and architectures, highlighting the amortized nature of these methods—allowing practitioners to bypass costly retraining when faced with new datasets. Neural Processes \cite{garnelo2018neural}, which emphasize inference over unstructured latent variables, have also benefited from the integration of attention mechanisms and transformer architectures \cite{kim2019attentive, nguyen2022transformer}.

In the context of dynamical systems, pretraining has proven effective for both time-series forecasting \cite{dooley2024forecastpfn} and the recovery of governing equations in physical systems \cite{dascoli2024odeformer}. These models are typically pretrained on large synthetic datasets, where the practitioner defines the family of processes to be recovered. Some approaches aim directly at predictive distributions \cite{ansari2024chronos}, while others attempt to infer the underlying generators. The latter has shown promise in modeling jump diffusion processes \cite{berghaus2024foundation} and stochastic differential equations \cite{seifner2025foundation}.

Our approach builds on these insights. Similar to \textit{prior-fitted networks}, we restrict the class of inferred systems by introducing a simulation-based prior that emphasizes compartment PK models, and places a stationary Ornstein–Uhlenbeck (OU) process on the parameters of the PK model. To account for the hierarchical structure of mixed-effects models, we incorporate unstructured latent variables for both fixed and random effects—drawing inspiration from neural processes—while retaining the flexibility of neural representations by not enforcing permutation invariance. In contrast to standard inference methods, which often assume fully observed trajectories, we focus on the posterior predictive distribution of partially observed systems. This formulation enables both, the generation of trajectories for unseen individuals and the forecasting for partially observed ones. Consequently, our model leverages the learned fixed and random effect representations in a unified generative and predictive framework.

\section{Preliminaries}

Synthetic data for model training are derived from a tailored combination of compartmental PK systems and stochastic (time-varying) parameters within these systems. 

\subsection{Compartmental Pharmacokinetic Systems}

Each simulated study is comprised of $I$ individuals, each characterized by a time-varying latent state of drug amount in gut, central (c) and peripheral (per) compartments
\[
\mathbf{X}(t)=
  \Bigl(
    X_{\text{gut}},\,
    X_{c},\,
    X_{\text{per},1},\dots,X_{\text{per},P}
  \Bigr)^{\!\top}\in\mathbb{R}^{2+P}
\]
which solves the ODE system
\begin{align}
\dot X_{\text{gut}}      &= -k_{a}\,X_{\text{gut}}, \nonumber\\
\dot X_{c}               &= k_{a}X_{\text{gut}}
                           -\Bigl(k_{e}+\sum_{j=1}^{P}k^{+}_{j}\Bigr)X_{c}
                           +\sum_{j=1}^{P}k^{-}_{j}X_{\text{per},j}, \label{eq:pk-odes}\\
\dot X_{\text{per},j}    &= k^{+}_{j}X_{c}-k^{-}_{j}X_{\text{per},j},
                           \qquad\qquad j=1,\dots,P. \nonumber
\end{align}
The initial condition is determined by the dose $u$ and the dosing route, i.e.
$X_{c}(0)=u$ for intravenous and $X_\text{gut}(0)=u$ for oral dosing (other states are initialized at 0).
The number of individuals and peripheral compartments, as well as the dose and dosing route, are randomly drawn for each simulated study.

\subsection{Stochastic parameter model}

Denote by $\theta_i(t)=\log ( k_a,  k_e,  V, k^+_1,  k^-_1,..., k^+_P, k^-_P) \in \mathbb{R}^{3+2P}$ the vector of log-kinetic parameters (plus volume of distribution $V$, which appears in the observation model below) for individual $i$. 
Each component $\theta_{i,k}$, $k\in\{1,\dots,3+2P\}=:\mathcal{K}$, evolves as an independent OU process
\begin{equation}
d\theta_{i,k}(t)=
  -\lambda_{k}\!\left(\theta_{i,k}(t)-\mu_{i,k}\right)\,dt
  +\sigma_{k}\,dW_{i,k}(t)
\label{eq:ou}
\end{equation}
initialized at $\theta_{i,k}(0)\sim \mathcal{N}(\mu_{i,k},\sigma^2/(2\lambda))$ to ensure stationarity, i.e., $\theta_{i,k}(t)\sim \mathcal{N}(\mu_{i,k},\sigma^2/(2\lambda))$ for $t\ge 0$.
The OU parameters $(\mu_{i,k},\sigma_{k},\lambda_{k})$ are themselves random, derived from the following independent samples:
\begin{center}
\begin{minipage}{0.48\linewidth}
\begin{align*}
m_k       &\sim \mathcal{U}(a_{m_k},b_{m_k}), \\
s_k       &\sim \mathcal{U}(a_{s_k},b_{s_k}), \\
\mu_{i,k}|m_k,s_k      &\sim \mathcal{N}(m_{k}, s_{k}^{2}),
\end{align*}
\end{minipage}
\hfill
\begin{minipage}{0.48\linewidth}
\begin{align*}
\lambda_{k}^2  &\sim \mathcal{U}(a_\lambda,b_\lambda), \\
\frac{\sigma_{k}^{2}}{2\lambda_k} & \sim \mathcal{U}(a_\sigma,b_\sigma).
\end{align*}
\end{minipage}
\end{center}
\vspace{0.1cm}
The values of $a_{m_k},b_{m_k},a_{s_k},b_{s_k}$ have been chosen based on a meta-analysis of NLME model fits; these values determine typical PK trajectories.
If we had $\theta_k(t)\equiv\mu_k$, the data-generating model would produce samples from a compartmental PK model. 
To allow for generalization beyond this parametric class, parameter stochasticity is induced via Eq.~\ref{eq:ou}. Its governing parameters $a_\lambda,b_\lambda,a_\sigma,b_\sigma$ have been determined empirically to produce plausible PK profiles. 

\subsection{Observation Model}

To derive meaningful sampling times per study, we first determine characteristic timescales of reaching peak plasma concentration ($T_\text{peak}$) and of drug half-life ($T_\text{half}$) based on a one-compartment model with first-order absorption:
\begin{equation*}
T_\text{peak} = \frac{\log(\bar k_a) - \log(\bar k_e)}{\bar k_a - \bar k_e},\qquad
T_\text{half} = \frac{\log(2)}{\bar k_e},
\end{equation*}
for typical parameters a the study level (i.e., $\bar k_a = e^{m_1}$ and $\bar k_e = e^{m_2}$).
Irregular plasma samples are then simulated at subject–specific times
\(0\le \tau_{i,1} < \dots < \tau_{i,T_i}\le \tau_\text{max}\). We define this specific times following the behavior of experimental study design. We select 4 equally spaced dense observations before $T_\text{peak}$, and 6 spaced observations after $T_\text{peak}$ at increasingly spaced times based on $T_\text{half}$, which mimics typical sampling schedules.
Finally, we randomly subsample the number of observations according to empirical sample size distributions, and additionally handle size mismatch through masking.
The final observations of perturbed plasma concentrations are then derived by selecting simulations at observations points and assuming a proportional measurement error:
\begin{equation}
Y^i_{j} \;=\; \frac{X^i_{c}(\tau^i_{j})}{V_{i}(\tau^i_{j})}
             (1\;+\; \varepsilon_{i,j}),
\quad
\varepsilon_{i,j}\sim\mathcal{N}\!\bigl(0,\sigma_{\text{obs}}^{2}\bigr).
\label{eq:obs}
\end{equation}

\subsection{Generative Data Model}

Collecting all elements and setting $\eta_k = (m_k,s_k,\lambda_k,\sigma_k)$ and $\boldsymbol{\eta}= (\eta_k)_{k\in\mathcal{K}}$, the joint probability used to generate a dataset
factorises as
\begin{align}
&p(\boldsymbol{\eta},T_\text{peak},T_\text{half},\boldsymbol{\theta},\boldsymbol{\mu},\boldsymbol{\tau},\mathbf{X},\mathbf{Y}) \nonumber \\[0.3em]
&=
  p_{\text{pop}}(\boldsymbol{\eta})\;
  p(T_\text{peak},T_\text{half}\mid\boldsymbol{\eta})\nonumber\\
&\quad\times\;
  \prod_{i=1}^{I}
  p_{\text{ind}}\!\left(\boldsymbol{\theta}_{i}(0),\boldsymbol{\mu}_{i}\mid\boldsymbol{\eta}\right) \, \times\nonumber\\[-0.8em]
&\qquad\quad\;\;\,\,
  p_{\text{dose}}\!\left(\boldsymbol{X}^{i}(0)\right) 
  p_{\text{time}}(\boldsymbol{\tau}^i\mid T_\text{peak},T_\text{half})
\nonumber\\[0.3em]
  &\quad\times\;
  \prod_{t\in\mathcal{T}}
     p_{\text{OU}}\!\left(\boldsymbol{\theta}_{i}(t+\Delta t)\mid\boldsymbol{\theta}_{i}(t),\boldsymbol{\eta},\boldsymbol{\mu}_{i}\right)\, \times\nonumber\\[-0.8em]
&\qquad\qquad
     p_{\text{ode}}\!\left(\mathbf{X}^{i}(t+\Delta t)\mid
     \mathbf{X}^{i}(t),\boldsymbol{\theta}_{i}(t)\right)
\nonumber\\[-0.2em]
&\quad\times\;
  \prod_{j=1}^{T_i}
     p_{\text{obs}}\!\left(Y^i_{j}\mid\mathbf{X}^{i}(\tau^i_{j}),\boldsymbol{\theta}_{i}(\tau^i_{j})\right),
\label{eq:pk-path-prob}
\end{align}
where $\mathcal{T}$ is a fine simulation grid,  
$p_{\text{OU}}$ denotes the OU transition density implied by \eqref{eq:ou},  
$p_{\text{ode}}$ is the flow of \eqref{eq:pk-odes}, and  
$p_{\text{obs}}$ is the observation model in \eqref{eq:obs}.  
Eq.~\eqref{eq:pk-path-prob} serves both, as a blueprint for synthetic data generation and as the conceptual backbone for our in–context learning objective.

\subsection{Problem Formulation}
Let $\mathcal{S}=\{\mathcal{D}^{i}\}_{i=1}^{I}$ denote the study
context, with data of individual~$i$ given by 
\(
\mathcal{D}^{i}=(\boldsymbol{u}^i, \boldsymbol{\tau}^{i},\mathbf{Y}^{i})
\), its observations times
$ \boldsymbol{\tau}^{i}= (\tau^{i}_{1},\dots,\tau^{i}_{T_i})$, observations 
$\mathbf{Y}^{i}= (y^{i}_{1},\dots,y^{i}_{T_i})$ and its dosing information (value and type) $\boldsymbol{u}^i$. Note that times need not be shared across individuals, as observation times are irregularly sampled. 

\paragraph{Function Discovery.}
Now we chose to specify our mixed-effects problem as a type of hierarchical function discovery, as each study defines family of functions $F^i:\tau \rightarrow Y^i| \mathcal{S}$. Here, $Y^i$ corresponds to the observation process that will be attained if one had access to all  hidden variables associated with individual $i$.  Similar to the neural process family, we now model a distribution over functions $G: \mathcal{S} \rightarrow P_{F^n|\mathcal{S}}(F^n:\tau_T\rightarrow Y^n_T)$.
The general problem of mixed effects, in essence, amounts to define a fixed effect that is associated to the specifics of the study, and random effects that are given per individual. 
Given a study context $\mathcal{S}$ produced by an \emph{unknown} instantiation of the generative model \eqref{eq:pk-path-prob}. At test time, we are given a \emph{new} subject with 
\(
\mathcal{D}^{n}=(\boldsymbol{u}^n,\boldsymbol{\tau}^{n},\mathbf{Y}^{n})
\),
and aim to forecast future concentrations or sample plausible
trajectories. 
We pursue two complementary goals: 
\begin{enumerate}
\item \textbf{Population synthesis.}  
      Without accessing any per–subject states, we should be able to generate new virtual individuals
      \(
      \bigl\{\mathcal{D}^{\,1},\dots,\mathcal{D}^{\,N}\bigr\}
      \)
      whose statistics are indistinguishable from those of $\mathcal{S}$.
      In other words, learn to \emph{sample} from the implicit population distribution. 
\item \textbf{Individual prediction.}  
      For a \emph{new} individual $n$, given zero, one, or a handful of early measurements
      \(\mathbf{Y}^{i^{\dagger}}_{1:J_0}
      \) as well as its corresponding dosing $\mathbf{u^n}$,
      forecast future concentrations
      \(Y^{n}(\tau^*)
      \) at any target time $\tau^*$.
\end{enumerate}
Crucially, our transformer learns these behaviors \emph{in‑context}: at deployment it receives $\mathcal{S}$ (and, for prediction, the partial record) as a prompt and returns samples or forecasts in a single forward pass, without gradient–based optimization.  
This design contrasts with classical NLME workflows that perform explicit inference for every new dataset.

\section{Model}
This section introduces the probabilistic model.
We endow our model with a hierarchy of latent variables: 
a \emph{global} study code (playing the role of the fixed effects), 
and an \emph{individual–specific} code (random effects) for every enrolled subject. 
We treat our model as a neural process but without permutational invariance. 
Here, the new individual is treated as a global target defined by $\mathcal{D}^{n}$, and the full data as $\mathcal{S}^{n}=\mathcal{S}\cup\mathcal{D}^{n}$ augments the context with the newcomer. Essentially, the model is defined via a loss on the global context and target (that we refer simply as new individual) and, subsequently, a context and target split for the individuals.

\begin{figure}[ht]
    \centering
    \includegraphics[width=0.65\linewidth]{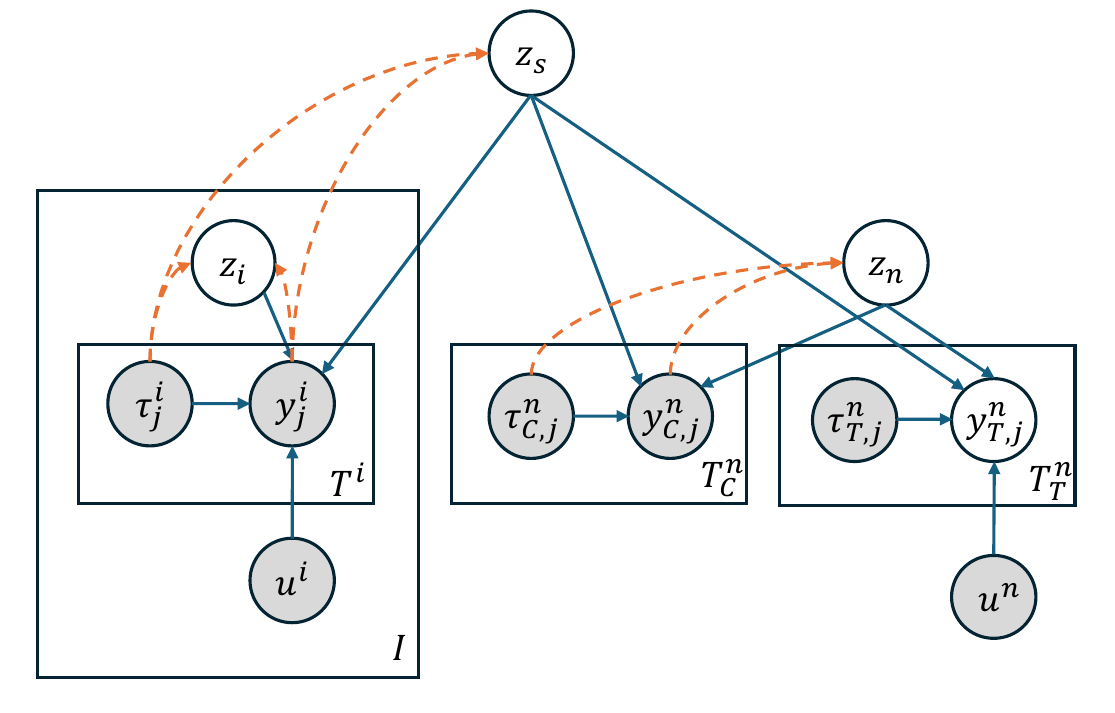}
    \caption{Hierarchical latent structure assumed in our AICMET model. 
        Shaded nodes are observed.
        All latent representations ($z_i,z_s,z_n$) are continuous; 
        solid blue arrows indicate conditional dependencies (decoder), orange dashed arrows indicate the recognition network (encoder).
        }
    \label{fig:architecture}
\end{figure}

\label{ssec:latent-generative}

\vspace{0.3em}\noindent\textbf{Latent codes.}
A zero–mean isotropic normal prior is placed on the \emph{global} study
vector $\,\mathbf{z}_{s}\in\mathbb{R}^{d_{s}}$ and on every
\emph{individual} vector $\,\mathbf{z}_{i}\in\mathbb{R}^{d_{i}}$,
including the unseen individual $n$:
\[
\mathbf{z}_{s}\sim\mathcal{N}(\mathbf{0},\mathbf{I}),\qquad
\mathbf{z}_{i}\stackrel{\text{i.i.d.}}{\sim}\mathcal{N}(\mathbf{0},\mathbf{I})
\quad\text{for }i\in\{1,\dots,I\}.
\]

\vspace{0.3em}\noindent\textbf{Likelihood.}
The joint density factorizes as
\begin{equation}
\begin{aligned}
p( \mathbf{Y} \mid \mathbf{z}_s, \{\mathbf{z}_i\}_{i=1}^I, \boldsymbol{\tau})
=\, \\
\times \prod_{i=1}^I \prod_{j=1}^{T_i}
\mathcal{N}\bigl(y^i_j \mid \mu_\theta(\tau^i_j, \mathbf{z}_s, \mathbf{z}_i),\, \sigma^2_\theta(\tau^i_j, \mathbf{z}_s, \mathbf{z}_i)\bigr)
\end{aligned}
\label{eq:decoder}
\end{equation}
\paragraph{Hierarchical Variational Posterior.} Exact posterior inference under \eqref{eq:decoder} is
intractable since $\mu_{\theta}$ and $\sigma_\theta$ are non–linear. We introduce the
factorized approximation
\begin{align}
q_\phi\bigl(\mathbf{z}_{s},\{\mathbf{z}_{i}\}_{i=1}^{I},\mathbf{z}_{n}
   \,\big|\,\mathcal{S}^{n}\bigr) \nonumber 
= \\ q_\phi(\mathbf{z}_{s}\mid\mathcal{S}^{n})\;
   q_\phi(\mathbf{z}_{n}\mid\mathcal{D}^{n}) 
\quad \times \prod_{i=1}^{I} q_\phi(\mathbf{z}_{i}\mid\mathcal{D}^{i}).
\label{eq:encoder}
\end{align}
Each factor is Gaussian with mean and diagonal
covariance output by an encoder network whose architecture will be described further below. We avoid mentioning $\phi$ 

\paragraph{New Individual Objective.} Our first objective is to maximize the \emph{predictive} likelihood of
the new individual, marginalizing all latent variables:
\(
\log p(\mathbf{Y}^{n}|\boldsymbol{\tau}^{\,n},\mathcal{S}).
\)
Applying Jensen’s inequality with the posterior
\eqref{eq:encoder} yields the evidence lower bound (ELBO):
\begin{equation}
\begin{aligned}
\mathcal{L}_{\text{new}}
&= \mathbb{E}_{q}
   \left[
     \log
     p\bigl(\mathbf{Y}^{n} \mid \boldsymbol{\tau}^{\,n},
            \mathbf{z}_{n}, \mathbf{z}_{s}\bigr)
   \right] \\
&\quad -\;
   \text{KL}\!\left[
     q(\mathbf{z}_{s} \mid \mathcal{S}^{n})
     \,\big\|\, q(\mathbf{z}_{s} \mid \mathcal{S})
   \right] \\
&\quad -\;
   \text{KL}\!\left[
     q(\mathbf{z}_{s} \mid \mathcal{S}^{n})
     \,\big\|\, p(\mathbf{z}_{s})
   \right] \\
&\quad -\;
   \text{KL}\!\left[
     q(\mathbf{z}_{n} \mid \mathcal{D}^{n})
     \,\big\|\, p(\mathbf{z}_{n})
   \right],
\end{aligned}
\label{eq:elbo-new}
\end{equation}
where the KL terms regularize the global and new–individual
encoders towards their priors. Different from a Variational Auto Encoders  the term $   \text{KL}\!\left[
     q(\mathbf{z}_{s} \mid \mathcal{S}^{n})
     \,\big\|\, q(\mathbf{z}_{s} \mid \mathcal{S})\right]$ ensures that the representation with and without the new individual remain the same, whereas the other KL terms allow for sampling as the structure code is similar to the prior. During training, the new individual is a randomly selected individual from the study that is kept outside the study context.
     
\paragraph{Predictive Objective.} For longitudinal prediction we split each \emph{observed} trajectory
into \emph{context} and \emph{target} subsets,
\(
\mathcal{D}^{i}
=\mathcal{D}^{i}_{\text{C}}\cup\mathcal{D}^{i}_{\text{T}},
\)
and denote the resulting corpus by
\(
\mathcal{S}_{\text{F}}=\{\mathcal{D}^{i}_{\text{C}}\cup
                          \mathcal{D}^{i}_{\text{T}}\}_{i=1}^{I}.
\)
Adapting \eqref{eq:elbo-new} to this partially observed setting gives
the forecast bound
\begin{equation}
\begin{aligned}
\mathcal{L}_{\text{F}}
 &= \mathbb{E}_{q(\mathbf{z}_{s}\mid\mathcal{S})}
    \Bigl[
      \sum_{i=1}^{I}\!
        \mathbb{E}_{q(\mathbf{z}_{i}\mid\mathcal{D}^{i}_{\text{F}})}
        \!\bigl[
          \log p(\mathbf{Y}^{i}_{\text{T}}\!\mid\boldsymbol{\tau}^{i}_{\text{T}},\mathbf{z}_{i},\mathbf{z}_{s})
        \bigr]
    \Bigr] \\
 &\quad - \sum_{i=1}^{I}
          \text{KL}\!\left[ q(\mathbf{z}_{i}\mid\mathcal{D}^{i}_{\text{F}})
          \,\middle\|\,
          q(\mathbf{z}_{i}\mid\mathcal{D}^{i}_{\text{C}}) \right],
\end{aligned}
\label{eq:forecast-elbo}
\end{equation}
where $\mathcal{D}^{i}_{\text{F}}=\mathcal{D}^{i}_{\text{C}}\cup
\mathcal{D}^{i}_{\text{T}}$. The KL term inside the sum encourages
the individual encoder to refine its beliefs after seeing the targets,
thereby specializing the latent to extrapolative prediction. Optimizing \eqref{eq:elbo-new} and \eqref{eq:forecast-elbo} furnishes a
study–level representation capable of (i) generating synthetic
individuals whose statistics match those of $\mathcal{S}$, and (ii)
forecasting future concentrations of any individual after observing a
handful of early samples. The neural parameterizations that realize
$q_\phi$, $\mu_{\theta}$ and $\sigma_{\theta}$, together with their training procedure, are the
subject of the next section.
\paragraph{Prediction.} For the predictive  distribution $d\mathbf{Z} = d\mathbf{z}_id\mathbf{z}_s$, we calculate the integral
\begin{equation}
p(\mathbf{Y}^{i}_{\text{T}}|\boldsymbol{\tau}^{i}_{\text{T}},\mathcal{S})
 =  \int p(\mathbf{Y}^{i}_{\text{T}}|\boldsymbol{\tau}^{i}_{\text{T}},\mathbf{z}_{i},\mathbf{z}_{s})q(\mathbf{z}_{s}|\mathcal{S})q(\mathbf{z}_{n}|\mathcal{D}^n_C)d\mathbf{Z}
\label{eq:predictive}
\end{equation}
via Monte Carlo by drawing samples from the encoders.

\section{Neural Network Architecture}

\begin{figure*}[t]
    \centering
    \includegraphics[width=0.75\linewidth]{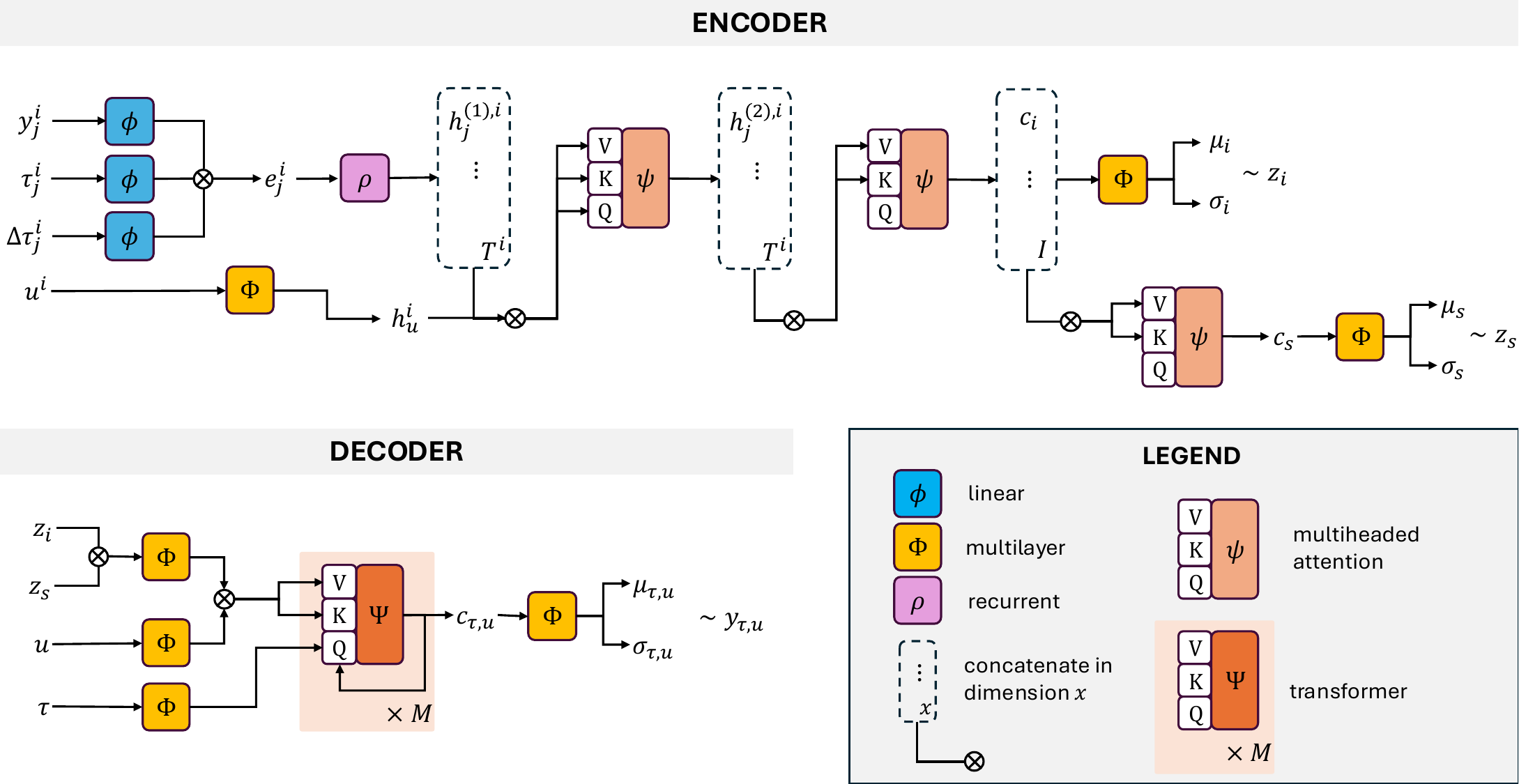}
    \vspace{-0.4em}
    \caption{The encoder produces dynamic representations with a recurrent backbone, and attention mechanisms are applied to summarize these representations at both the individual and study levels. Our transformer-based decoder embeds the encoder representations alongside dose information. Finally, we define functional queries that allow us to evaluate the predictive distribution at any target time $\tau$. By introducing $\mathbf{z}_i$ and $\mathbf{z}_s$, we can model fixed and random effects, enabling a population-aware, individualized characterization of dynamics.}
    \label{fig:architecture}
\end{figure*}

\newcommand{\btau}{\boldsymbol{\tau}}
\newcommand{\bY}{\mathbf{Y}}
\newcommand{\bz}{\boldsymbol{z}}

We now introduce the neural network architecture details for the Amortized In Context Mixed Effect Transformer (AICMET) model, required to define both encoder \eqref{eq:encoder} and decoder \eqref{eq:decoder}. We can see that in order to create a faithful representation of the data, we first need to create a longitudinal representation per patient that respects the time information per individual. 
Then, we aggregate the time dimension to obtain a representation per individual, and finally we aggregate all the individuals to obtain one representation per study. 
In the following, we use  \(H\) as hidden dimension and \(Z_{d}\) as latent dimension for the unstructured latent variables. A superscript in parentheses, e.g.\ \(\mathbf{h}^{(1)}\), denotes the
\emph{layer index}.
 \begin{figure*}[t]
    \centering
    \begin{minipage}[b]{0.31\textwidth}
        \centering
        (a)
        \adjustbox{trim=0pt 0pt 0pt 0pt, clip}{
            \includegraphics[width=\linewidth]{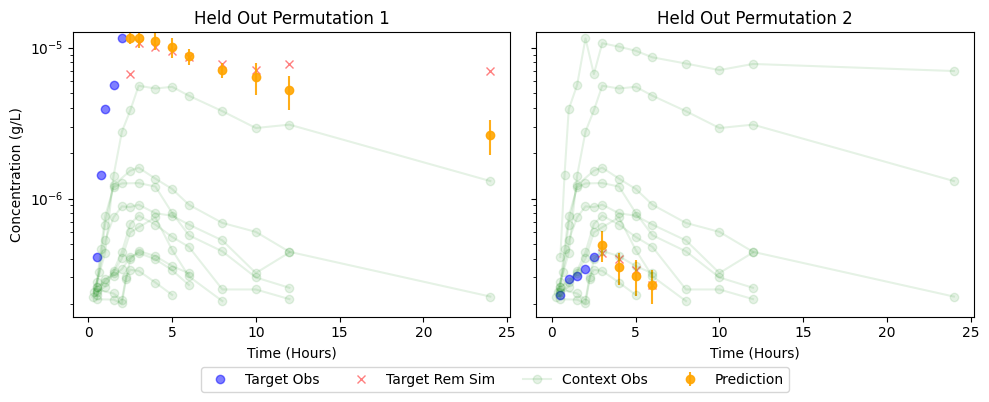}
        }
    \end{minipage}
    \hspace*{0.015\textwidth}
    \begin{minipage}[b]{0.31\textwidth}
        \centering
        (b)
        \adjustbox{trim=0pt 0pt 0pt 0pt, clip}{
            \includegraphics[width=\linewidth]{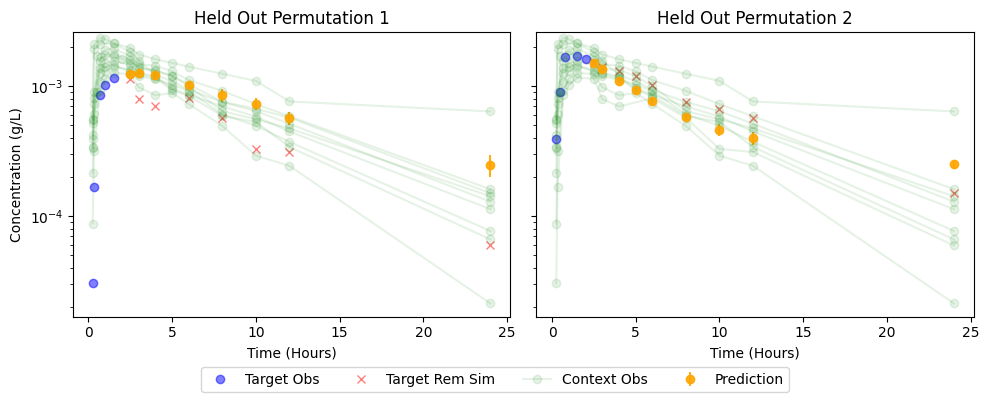}
        }
    \end{minipage}
    \hspace*{-0.04\textwidth}
    \begin{minipage}[b]{0.31\textwidth}
        \centering
        \hspace{1cm}(c)
        \adjustbox{trim=0pt 0pt 0pt 0pt, clip}{
            \includegraphics[width=0.71\linewidth]{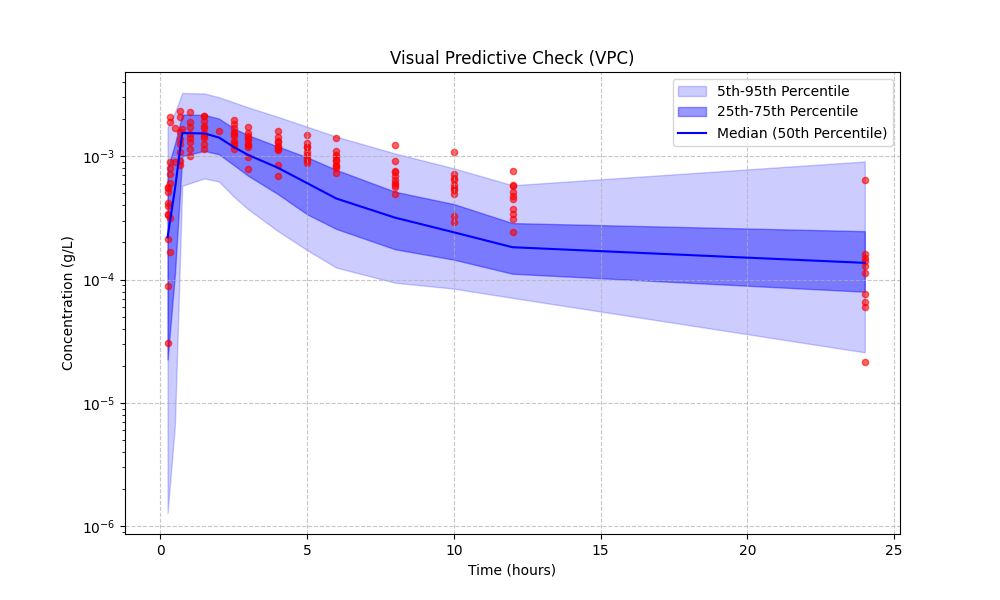}
        }
    \end{minipage}

    \vspace{-0.5em}

    \begin{minipage}[b]{0.31\textwidth}
        \centering
        \adjustbox{trim=0pt 0pt 0pt 0pt, clip}{
    \includegraphics[width=\linewidth]{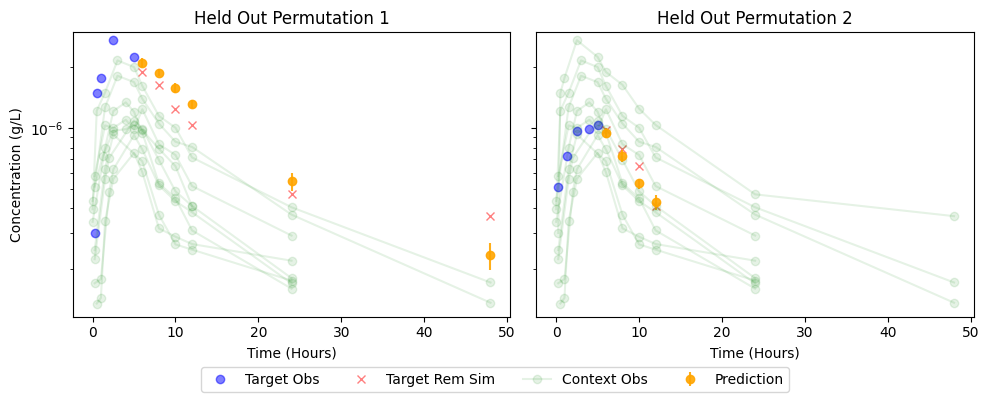}
        }
        (d)
    \end{minipage}
    \hspace*{0.015\textwidth}
    \begin{minipage}[b]{0.31\textwidth}
        \centering
        \adjustbox{trim=0pt 0pt 0pt 0pt, clip}{
            \includegraphics[width=\linewidth]{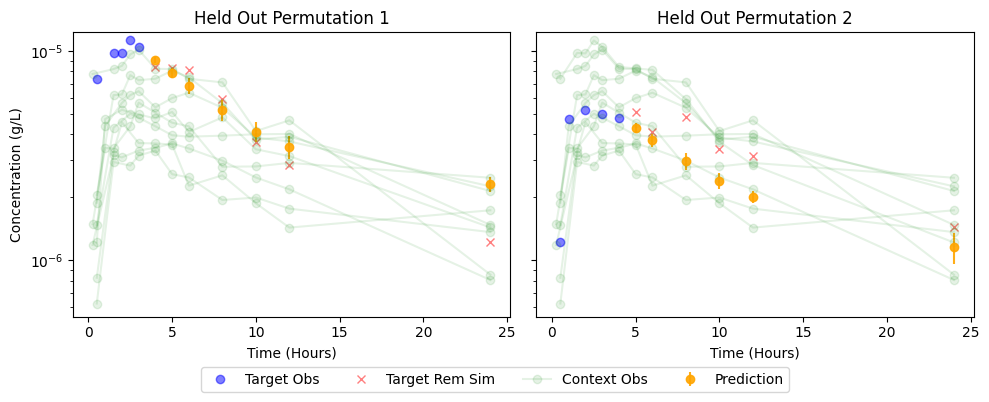}
        }
        (e)
    \end{minipage}
    \hspace*{-0.04\textwidth}
    \begin{minipage}[b]{0.31\textwidth}
        \centering
        \adjustbox{trim=0pt 0pt 0pt 0pt, clip}{
        \includegraphics[width=0.71\linewidth]{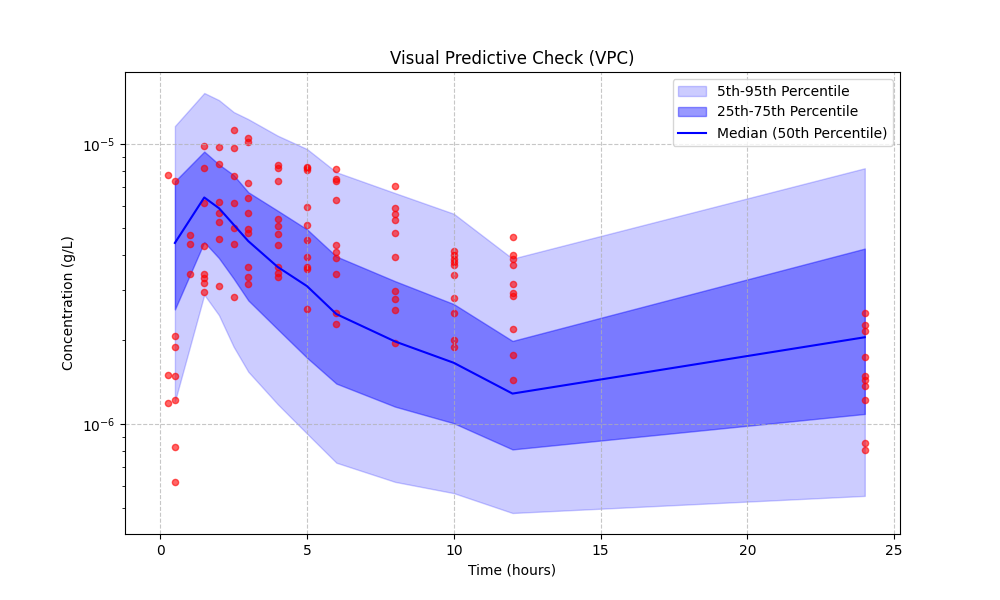}
        }
        \hspace{1cm}(f)
    \end{minipage}
    \caption{Predictive plots---(a) dextromethorphan, (b) caffeine, (d) rosuvastatin (e) 4-hydroxytolbutamide---and visual predictive checks---(c) caffeine and (f) 4-hydroxytolbutamide---for different compounds. Each subplot shows observed concentrations and simulation-derived prediction intervals.}
    \label{fig:vpc_grid}
\end{figure*}

\vspace{4pt}
\noindent\textbf{Neural-network primitives.} We denote by $\phi^{\theta}(\cdot)$ a linear projection $\mathbb{R}^{m} \to \mathbb{R}^{H}$, by $\Phi^{\theta}(\cdot)$ a feed-forward network $\mathbb{R}^{H} \to \mathbb{R}^{H}$, by $\rho^{\theta}(\cdot)$ a recurrent layer (GRU/LSTM), by $\psi^{\theta}(\mathbf{q},\mathbf{K},\mathbf{V})$ a multi-head attention module, and by $\Psi^{\theta}(\mathbf{Q},\mathbf{K},\mathbf{V})$ a transformer encoder (linear attention).
For both $\psi^\theta(\cdot, \cdot, \cdot)$ and $\Psi^\theta(\cdot, \cdot, \cdot)$ denote transformer encoders with linear attention ~\citep{katharopoulos2020transformers}, both of which take three arguments as inputs (i.e., queries, keys and values). 
We use the dot-product softmax kernel, i.e.,
\(
\psi^\theta(\mathbf{Q}, \mathbf{K}, \mathbf{V}) := \operatorname{softmax}\left( \frac{\mathbf{Q} \mathbf{K}^\top}{\sqrt{d_k}} \right) \mathbf{V}.
\)
Since PK data are often sparsely and irregularly sampled, we expand the features by defining the individual dataset as
\(
\widetilde{\mathcal{D}}^{i}
  =\bigl\{(y_{j},\tau_{j},\Delta\tau_{j})\bigr\}_{j=1}^{T_{i}-1}$ wiht $\Delta\tau_{j}=\tau^{i}_{j+1}-\tau^{i}_{j}.
\)
This simplifies the task of the network of handling irregular time intervals, as this representation enforces this feature.
\subsection{Encoder}
We first introduce the encoder, which we construct by first introducing the individual representation and then aggregating for the studies representations. 
\paragraph{Per-transition embeddings.}
We pass the basic features trough a linear layer and concatenate, thereby defining a basic data embedding: 
\[
\mathbf{e}^{(0),i}_{j}
 =\text{concat}\Bigl[
     \phi^{\theta}_{y}(y_{j}),
     \phi^{\theta}_{\tau}(\tau_{j}),
     \phi^{\theta}_{\Delta\tau}(\Delta\tau_{j})
   \Bigr]\in\mathbb{R}^{H}.
\]
\paragraph{Longitudinal Representation.} To capture the dynamical information from the embeddings, we now create a representation per individual and per time step, using either a recurrent neural network (GRU/LSTM) or a transformer architecture:
\[
\mathbf{h}^{(1),i}_{j}
=\rho^{\theta}\bigl(\mathbf{e}^{(0),i}_{1:j}\bigr)\in\mathbb{R}^{H}.
\]
 Next, in order to judge the overall curve shape,  a self-attention mechanism is applied over the longitudinal representations $\mathbf{h}^{(1),i}_{j}$. 
 This allows the different native features of the drug profiles, such as the peak and slope of different sections around the peak, to be weighted and compared against each other. We first stack the representations
\(
\mathbf{H}^{(1),i}=[\mathbf{h}^{(1),i}_{1},\dots,\mathbf{h}^{(1),i}_{T_{i}-1}]
 \in\mathbb{R}^{H\times(T_{i}-1)}
\); then
\[
\mathbf{H}^{(2),i}
 =\Psi^{\theta}\!\bigl(\mathbf{H}^{(1),i},
                       \mathbf{H}^{(1),i},
                       \mathbf{H}^{(1),i}\bigr).
\]
\paragraph{Individual posterior.}
Now we aggregate with attention pooling so that we are able to obtain a representation per individual
\[
\mathbf{c}_{i}
 =\psi^{\theta}\!\bigl(\mathbf{q},
                       \mathbf{H}^{(2),i},
                       \mathbf{H}^{(2),i}\bigr)\in\mathbb{R}^{H}.
\]
We then differentiate the mean and variance of the encoder with a final MLP layer
\[
q(\bz_{i}\mid\mathcal{D}^{i})
 =\mathcal{N}\!\Bigl(
     \bz_{i}\in\mathbb{R}^{Z_{d}}
     \;\bigl|\;
     \Phi^{\theta}_{\mu}(\mathbf{c}_{i}),
     \operatorname{diag}\!\bigl(
       e^{\,\Phi^{\theta}_{\sigma}(\mathbf{c}_{i})}
     \bigr)
   \Bigr).
\]
\paragraph{Study Posterior.}
To obtain a study-specific representation, we concatenate the individual summaries
\(
\mathbf{C}=[\mathbf{c}_{1},\dots,\mathbf{c}_{I}]
 \in\mathbb{R}^{H\times I},
\)
and then summarize with another attention pooling mechanism,
\(
\mathbf{c}_{s}
  = \psi^{\theta}\bigl(\mathbf{q}_{s}, \mathbf{C}, \mathbf{C}\bigr).
  \)
Finally the study encoder is obtained by
\[
q(\bz_{s} \mid \mathcal{S})
  = \mathcal{N}\!\Bigl(
     \bz_{s} \in \mathbb{R}^{Z_{d}} \;\bigl|\;
     \Phi^{\theta}_{\mu}(\mathbf{c}_{s}),
     \operatorname{diag}\!\bigl(
       e^{\,\Phi^{\theta}_{\sigma}(\mathbf{c}_{s})}
     \bigr)
   \Bigr).
\]

\subsection{Decoder}
\vspace{-0.08cm}
To design a decoder suited for PK data, we must accommodate the irregular sampling schedules typically observed in clinical datasets while simultaneously exploiting the strong \emph{context-learning} capabilities of modern transformers. Conventional neural ODE approaches (e.g., \cite{lu2021deep}) rely on adjoint-sensitivity computations; this can hinder scalability for a data prior like ours that requires many samples to enforce expert based inductive bias. 
We therefore adopt a transformer decoder endowed with \emph{functional attention}~\cite{seifner2025foundation}. The requested prediction/sample time points $\tau$ are embedded as \textbf{queries}, whereas the dosing information $\mathbf{u}^n$ together with the population-level ($\mathbf{z}_{s}$) and individual-level ($\mathbf{z}_{n}$) latent variables are embedded as \textbf{keys} and \textbf{values}. Through self-attention, the model acts as a context learner: each query time dynamically attends to the most informative dosing and latent-effect context, thereby defining a distribution over concentration-time functions conditioned on both, $\mathbf{z}_{s}$ and $\mathbf{z}_{i}$. For a new subject $\mathcal{D}^{(n)}$, we first sample $\mathbf{z}_{n}$ from its prior, then propagate the query-time set through the decoder to obtain the predictive concentration distribution at each requested time point.
Specifically, for a new individual \(\mathcal{D}^{n}\), we first sample
\(
\bz_{n}\sim q(\bz_{n}|\mathcal{D}^{n})\) and \(\bz_{s}\sim q(\bz_{s}|\mathcal{S})
\) for prediction, or \(\bz_n \sim \mathcal{N}(0,I)\) for generation
and embed a query time \(\tau\): 
\begin{align}
\mathbf{q}_{\tau} 
  &= \phi^{\theta}_{\text{dec}}(\tau) \in \mathbb{R}^{H}, \nonumber \\
\mathbf{K}_{n} 
  &= \Phi^{\theta}_{K}\left([\bz_{n}; \bz_{s}; \mathbf{u}]\right) \in \mathbb{R}^{H \times 1}, \nonumber  \\
\mathbf{V}_{n} 
  &= \Phi^{\theta}_{V}\left([\bz_{n}; \bz_{s}; \mathbf{u}]\right) \in \mathbb{R}^{H \times 1}, \nonumber  \\
\mathbf{c}_{\tau}
  &= \psi^{\theta}\left(\mathbf{q}_{\tau}, \mathbf{K}_{n}, \mathbf{V}_{n}\right) \in \mathbb{R}^{H}.
\end{align}
It only remains to define the final mean and variance heads,
\(
(\mu_{\tau},\log\sigma^{2}_{\tau})
 =\Phi^{\theta}_{\text{dec}}(\mathbf{c}_{\tau}),
\) and specify the distribution to a diagonal Gaussian 
\(
p\bigl(y_{\tau}\mid\tau,\bz_{n},\bz_{s}\bigr)
=\mathcal{N}\bigl(y_{\tau}\mid\mu_{\tau},\sigma^{2}_{\tau}\bigr).
\)

\section{Results}

\begin{table*}[t]
    \centering
    \renewcommand{\arraystretch}{1.2} 
    \rowcolors{3}{orange!15}{white} 
    \setlength{\tabcolsep}{5pt} 
    \resizebox{0.7\linewidth}{!}{ 
\begin{tabular}{lrrrrrrr}
\toprule
Compound & NLME & NODE-PK & T-PK &  SNODE-PK & ST-PK & AICME-RNN & AICMET \\
\midrule
caffeine & \textbf{0.356} & 0.914 & 0.575 & 0.780 & 0.984 & 0.646 & 0.477 \\
dextromethorphan & 0.796 & 0.668 & 0.630 & 1.702 & 1.412 & 0.640 & \textbf{0.437} \\
digoxin & \textbf{0.315} & 1.403 & 0.717 & 0.501 & 0.421 & 0.569 & 0.457 \\
memantine & 0.411 & 0.549 & 0.799 & 0.580 & 0.869 & 0.534 & \textbf{0.362} \\
midazolam & 0.674 & 0.456 & 0.735 & 0.874 & 0.817 & 0.548 & \textbf{0.366} \\
omeprazole & 1.470 & 1.940 & 1.864 & 1.267 & 1.078 & 1.395 & \textbf{1.139} \\
paracetamol & \textbf{0.319} & 1.094 & 0.825 & 1.115 & 1.050 & 0.691 & 0.406 \\
repaglinide & 0.632 & 0.879 & 0.846 & 1.514 & 1.246 & \textbf{0.562} & 0.583 \\
rosuvastatin & 0.470 & 0.471 & 0.748 & 0.624 & 0.604 & 0.578 & \textbf{0.396} \\
tolbutamide & 0.766 & \textbf{0.683} & 0.816 & 0.949 & 0.998 & 0.854 & 0.691 \\
1-hydroxy-midazolam & -- & 0.741 & 0.678 & 1.395 & 1.216 & 0.935 & \textbf{0.729} \\
4-hydroxy-tolbutamide & -- & 0.871 & 0.898 & 0.524 & 0.742 & 0.274 & \textbf{0.265} \\
5-hydroxy-omeprazole & -- & 2.014 & 1.683 & 1.811 & 1.600 & \textbf{1.575} & 1.615 \\
dextrorphan & -- & 0.723 & 1.001 & 0.904 & 0.860 & 0.614 & \textbf{0.374} \\
hydroxy-repaglinide & -- & 0.340 & 0.532 & 0.059 & 0.336 & \textbf{0.095} & 0.113 \\
omeprazole sulfone & -- & 1.992 & 1.620 & 1.529 & \textbf{1.294} & 1.438 & 1.366 \\
paracetamol glucuronide & -- & 0.509 & 0.423 & 0.823 & 1.057 & 0.365 & \textbf{0.295} \\
paraxanthine & -- & 1.648 & 0.646 & 0.653 & 0.858 & 0.409 & \textbf{0.266} \\
\bottomrule
\end{tabular}
    }
    \caption{\textbf{Comparison of log-RMSE across models.} For each compound, log-RMSE is reported for baseline models (NLME and NODE-PK), and the newly proposed AICMET model including its ablations. The best-performing model for each compound is highlighted in \textbf{bold}.}
    \label{tab:main_results}
\end{table*}

\subsection{Datasets}
Evaluation data for performance of the AICMET model were extracted from the open-access database PK-DB, which contains data from both clinical trials and pre-clinical studies \cite{grzegorzewski2021pkdb}. 
We obtained the plasma concentration measurements for 18 different compounds (totalling  2019 valid concentration–time data points) from a Phase I clinical trial designed to establish the safety and appropriate dosing of a combination of drugs and their metabolites in healthy volunteers \cite{lenuzza2016safety}. 
The curated dataset includes 10 parent compounds 
(caffeine, dextromethorphan, digoxin, memantine, midazolam, omeprazole, paracetamol, repaglinide, rosuvastatin and tolbutamide) 
and their primary metabolites 
(paraxanthine, dextrorphan, 1-hydroxy-midazolam, 5-hydroxy-omeprazole, hydroxy-repaglinide, omeprazole sulfone, paracetamol glucuronide and 4-hydroxy-tolbutamide). 
Notably, the intra- and inter-substance sampling frequencies varied significantly, leading to irregularly sampled time series across both, subjects and compounds. There were 10 individuals per compounds and a maximum of 15 observations per individual.

\subsection{Baselines} We consider two baseline approaches for our newly developed AICMET framework: \emph{nonlinear mixed-effect modelling}, the standard in PK modelling, and \emph{neural ODEs}.
\paragraph{Nonlinear Mixed-Effect (NLME) Modelling.} For each of the 10 parent compounds in the evaluation dataset, one- and two-compartment NLME models were fitted to the dataset (with first-order oral absorption in case of orally dosed compounds), with inter-individual variability on all model parameters and a combined additive/proportional error model.
Subsequently, the better fitting model of the two was selected according to AIC and used as baseline model for that compound.
Since the production rates of metabolites (i.e., the ''dosing'' information) are depend on the PK of the parent compound, they could not be modelled this way. 
\paragraph{Neural ODEs (NODE-PK).} The neural ODE model for pharmacokinetic modelling defined in \cite{lu2021deep} is a dosing-aware extension of the latent neural ODE proposed in \cite{rubanova2019latent}, where neural ODE steps are used to evolve in between observations of a RNN. Crucially, we train this model in our simulated data.

\paragraph{Ablations.} To verify the effectiveness of our modelling approach, we construct a series of models by incorporating the different elements of our proposed representations, building on top of \textbf{NODE-PK} architecture. \textbf{T-PK}: our decoder trained on simulated data, but only on the prediction loss $\mathcal{L}_F$ without the KL terms. \textbf{SNODE-PK}: aggregates the encoder representation with attentions within the study before using the \textbf{NODE-PK} decoder. \textbf{ST-PK}: does the same but with our transformer architecture. \textbf{AICME-RNN} and \textbf{AICMET}: corresponds to our proposed methodology, where all losses are included, but \textbf{AICME-RNN} uses \textbf{NODE-PK} as decoder, whereas \textbf{AICMET} uses our transformer.

\paragraph{Evaluation Metrics.} For each compound, individual predictions were derived from PK data on all other individuals in the study plus the first 4 PK samples of the considered individual.
Prediction accuracy on the remaining PK samples was then assessed as root mean squared error on log-concentration scale (log-RMSE).
To assess the generative capabilities of AICMET model, we performed a simulation-based graphical evaluation called \emph{visual predictive check} (VPC), which is well-established in the PK community \cite{holford2005}.
This diagnostic tool compares the variability of simulated plasma concentration profiles within a study via an overlay of simulation percentiles and data. 

\paragraph{Implementation Details.} We used the ADAM optimizer with a learning rate of $0.0001$ and a batch size of $128$ across all datasets. 
Data were simulated on the fly and models were trained up until $5000$ iterations. The history context length of the decoder $|\mathcal{D}^n_C|$ varies according to the distribution of the empirical data. All experiments are conducted on a single Nvidia V-100 GPU, and results are based on $5$ runs per model. We rescaled the total loss with $ \mathcal{L}_T =   \sum_l e^{-U^\theta_l} \mathcal{L}^\theta_l(\mathcal{D})  -U^\theta_l$ as specified by
\citet{karras2024analyzing}, where $\mathcal{L}^\theta_l$ specify all the different elements of the loss (KL terms treated separately), and $U^\theta_l$ are nuisance parameters whose purpose is to ensure equal loss contributions of each component during training. Details of the architecture, training and simulations are in the Supplementary Material.
\subsection{Discussion}
As shown in Table \ref{tab:main_results}, AICMET achieves state-of-the-art performance, improving on the predictive power of NLME models for 7 out of 10 of parent compounds. In contrast to NLME models, our methodology is able to generalize to metabolites without any extra model specification! 
It also outperforms the pure neural ODE approach for all but one compound. Figure \ref{fig:vpc_grid} shows that the model accurately predicts across different scales as expected for PK data, and VPCs demonstrate that it captures the distribution of samples in new individuals. Crucially, our models is the first to provide an in context solution that generalizes across compounds. Different from other models, our methodology is able to outperform with zero-shot inference. NLME models were trained on data and required detailed hyperparameter tuning per compound, whereas our method is able to infer in seconds without tuning, and with better performance. Important future work will be to extend the proposed framework to handle multiple dosing and individual-specific covariates. We hypothesis that for the latter case, zero-shot inference may be insufficient and finetuning required. Moreover, better sampling methods can be included for the decoder as diffusion models  can handle more complex distributions.



\section{Acknowledgments}
The research has been partially funded by the Deutsche Forschungsgemeinschaft (DFG) – Project-ID 318763901 – SFB1294.

\bibliography{aaai2026}

\end{document}